\documentclass{esannV2}
\usepackage[dvips]{graphicx} 
\usepackage[utf8]{inputenc}
\usepackage{amssymb,amsmath,array}

\usepackage[misc]{ifsym}

\usepackage[utf8]{inputenc} 
\usepackage{hyperref}       
\usepackage{url}            
\usepackage{booktabs, multirow}       
\usepackage{amsfonts}       
\usepackage{nicefrac}       
\usepackage{microtype}      

\begin{document}
\title{Recurrent Feedback Improves Recognition of Partially Occluded Objects}

\author{Markus Roland Ernst$^{1,2}$\textsuperscript{(\Letter)}, Jochen Triesch$^{1,2}$, and Thomas Burwick$^{1,2}$
%
\thanks{This work was supported by the European Union’s Horizon 2020 research and innovation programme under grant agreement N\textsuperscript{\underline{o}} 713010  (GOAL-Robots, Goal-based Open-ended Autonomous Learning Robots).}
%
\vspace{.3cm}\\
%
%
1- Frankfurt Institute for Advanced Studies \\ 
Ruth-Moufang Stra\ss e 1, 60438 Frankfurt am Main - Germany
%
\vspace{.1cm}\\
2- Goethe-Universit\"at Frankfurt \\ 
Max-von-Laue-Stra\ss e 1, 60438 Frankfurt am Main - Germany\\
}

\maketitle
\begin{abstract}
Recurrent connectivity in the visual cortex is believed to aid object recognition for challenging conditions such as occlusion. Here we investigate if and how artificial neural networks also benefit from recurrence. We compare architectures composed of bottom-up, lateral and top-down connections and evaluate their performance using two novel stereoscopic occluded object datasets. We find that classification accuracy is significantly higher for recurrent models when compared to feedforward models of matched parametric complexity. Additionally we show that for challenging stimuli, the recurrent feedback is able to correctly revise the initial feedforward guess. 
\end{abstract}

\section{Introduction}
Primate object recognition has been widely assumed to be a feedforward process \cite{dicarlo2012visualobject}. This view is supported by the primate visual system's ability to accomplish the task within a mere 150~ms and by the recent success of feedforward convolutional neural networks \cite{potter1976shortterm, krizhevsky2012imagenet}. However, both neurobiological and computational evidence hint at the importance of recurrent connectivity \cite{kar2019evidence}. In particular, for the recognition of degraded or occluded objects, delayed behavioural and neural responses have been observed, which would allow for competitive processing via lateral recurrent connections \cite{johnson2005recognition}. For occluded stimuli, \cite{tang2014spatiotemporal} suggest that recurrent top-down connections may reconstruct missing information.
Whether object recognition in artificial neural networks can benefit from recurrent connections is less clear, however. Early investigations of this question used highly restricted datasets, where artificial inputs were partly faded out or masked \cite{spoerer2017recurrent, oreilly2013recurrent}. Under natural conditions, however, occlusion is highly dependent on viewing angle and primates perceive objects stereoscopically with two eyes. Thus, building on our previous work  \cite{ernst2019discounting}, we here developed two novel occluded image datasets that capture the full range of disparity and perspective cues for both natural (handwritten digits) and computer rendered (full 3-D objects) stimuli.
Using these datasets we compare a range of parameter-matched convolutional neural network models and demonstrate significant performance advantages for models that include recurrent connections.

\section{Methods}
\begin{figure}[hbtp]
\centering
\includegraphics[width=0.49\textwidth]{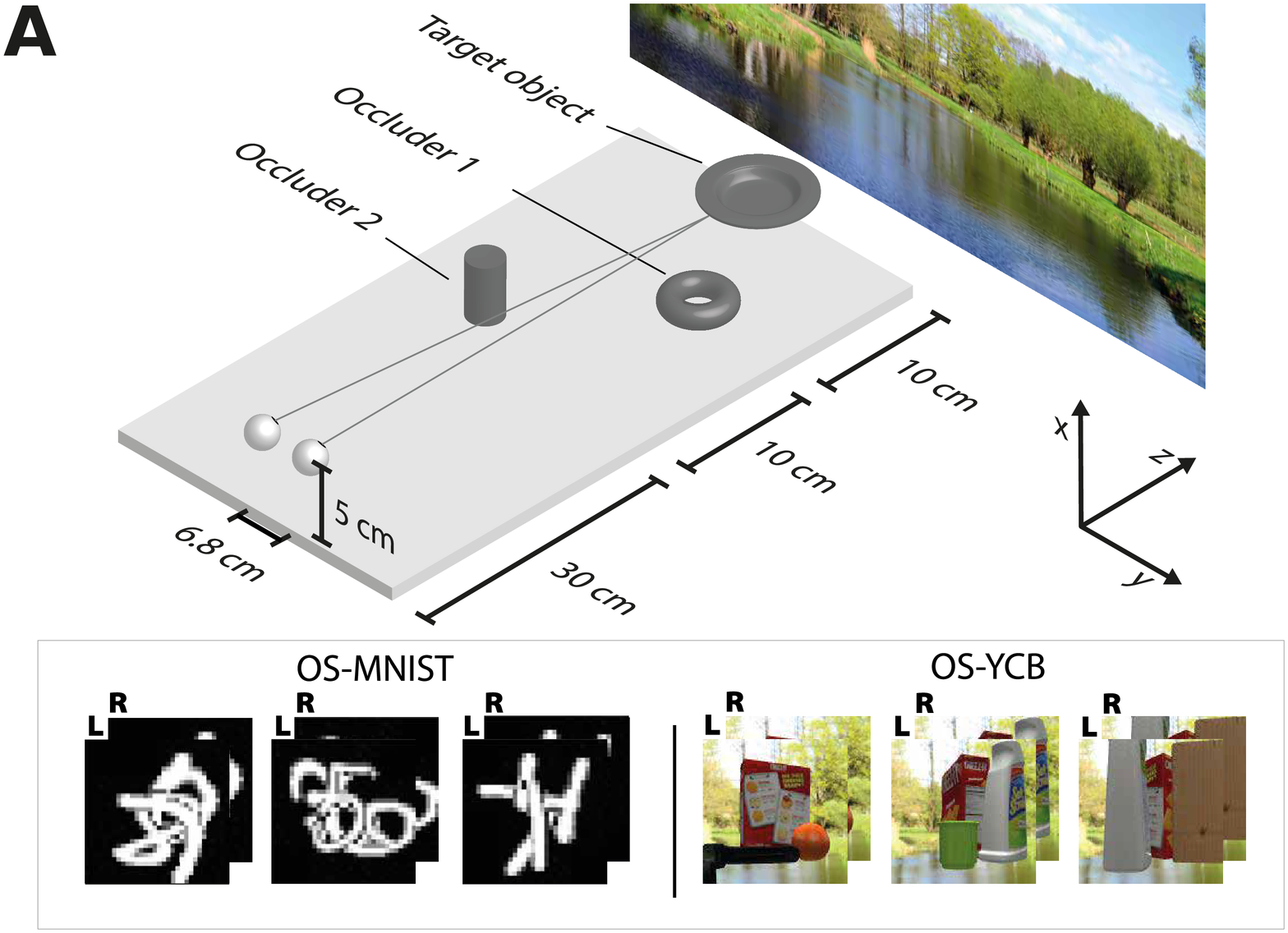}
\includegraphics[width=0.49\textwidth]{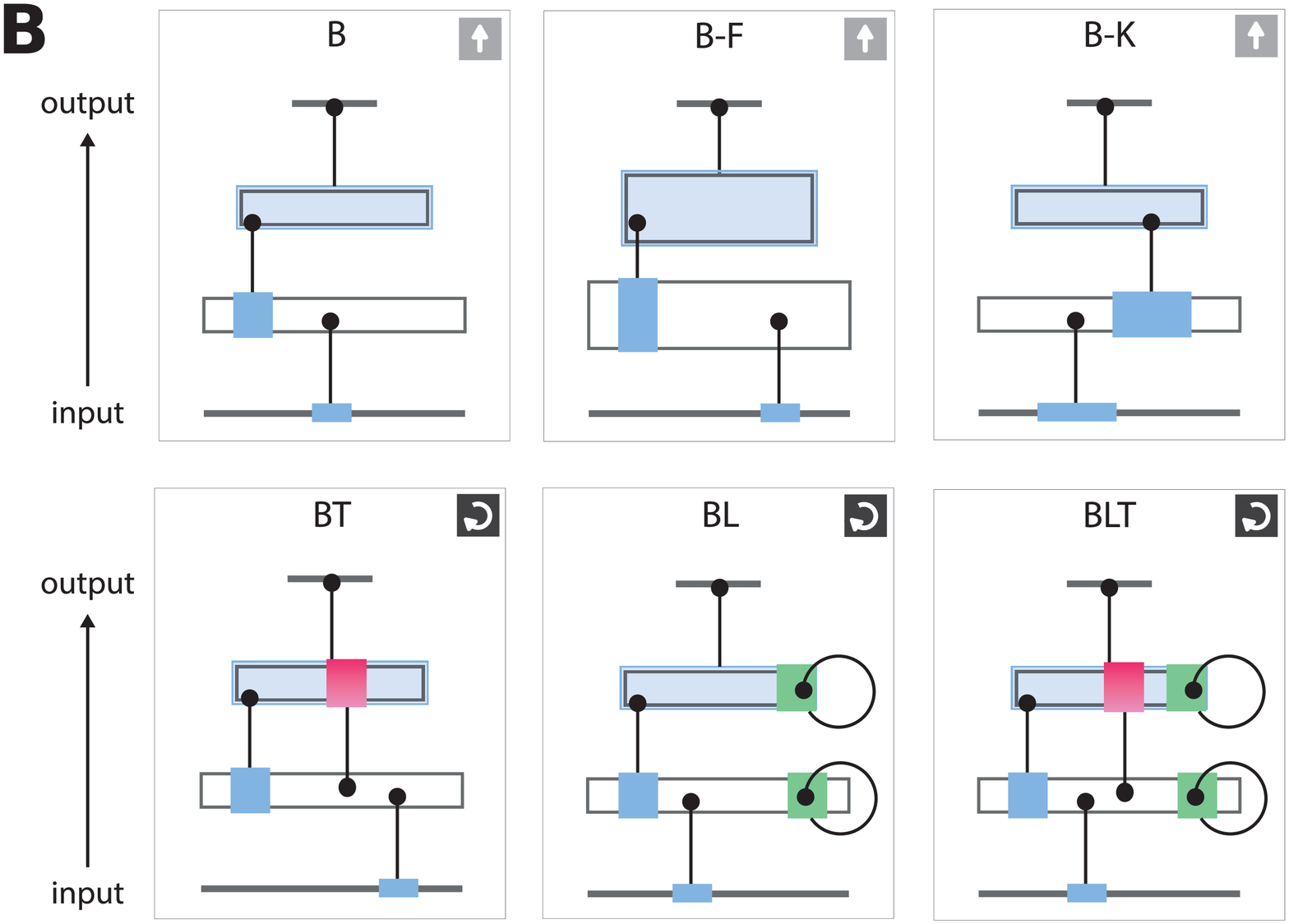}
\caption{Data sets and network models. (A) The centered target object is occluded by 2 additional objects arranged in a 3D-fashion. (B) A sketch of the six models named after their connection properties. B: bottom-up, L: lateral, T: top-down.}
\label{fig:network_overview}
\end{figure}

\paragraph{Occluded Stereo Multi-MNIST (OS-MNIST)}
Our first dataset is based on the MNIST digits \cite{lecun1998gradient}. It forces the networks to learn a representation that generalizes to different variants of a particular class.
Contrary to past studies \cite{oreilly2013recurrent, tang2014spatiotemporal, wyatte2012limits}, occlusion is generated by overlaying the target digit with two other digit instances in a pseudo-3D environment as shown in Fig.~\ref{fig:network_overview} A. Target and occluders are randomly distributed along the x,y-plane.
For each MNIST digit we generated 10 random occluder combinations, resulting in a total of $600{,}000$ stereo image pairs for training and $100{,}000$ for testing.
\paragraph{Occluded Stereo YCB-Objects (OS-YCB)}
Our OS-YCB data set contains stereo image pairs of 79 common household objects occluding one another \cite{calli2015benchmarking}. For each image, we placed three virtual 3D objects according to Fig.~\ref{fig:network_overview} A. The target-object is centered, occluders are randomly distributed along the y-axis. All objects are placed in upright position and turned by a random yaw angle. A background was chosen to simulate a context with natural image statistics. We generated $4{,}000$ images per object resulting in $316{,}000$ stereo image pairs, split \nicefrac{80}{20} for training and testing.
\\\\
All stimuli were rendered at $32 \times 32$ pixels, occluders were chosen in a way that no two instances of one class would appear in the same image, and occlusion was constrained to range between $20$ and $80\%$. For stereo input, the target-object is presented at zero disparity. Both datasets are available online.\footnote{\href{https://doi.org/10.5281/zenodo.3540900}{https://doi.org/10.5281/zenodo.3540900}}

Each model considered consists of an input layer, two hidden layers and an output layer. The two hidden layers allow for four basic models (\emph{B}, \emph{BT}, \emph{BL}, \emph{BLT}). Bottom-up and lateral connections are implemented as convolutional layers with a stride of $1 \times 1$ followed by a $2 \times 2$ maxpooling operation with a stride of $2 \times 2$.
Top-down connections are realized as $3 \times 3$ transposed convolutions \cite{zeiler2010deconvolutional} with output stride $2 \times 2$. Each of the recurrent network models is unrolled for four time steps and trained by backpropagation through time \cite{rumelhart1986learning}. 
 When reporting accuracy, we consider the output at the last unrolled time step.
To compensate for the increased parameter space of recurrent models, we introduce two additional feedforward models \emph{B-F} and \emph{B-K} as in \cite{spoerer2017recurrent}. \emph{B-F} doubles the number of convolutional filters per layer from 32 to 64, \emph{B-K} has larger $5 \times 5$ kernels compared to $3 \times 3$ in all other models.

\paragraph{Layers}
The inputs to the hidden recurrent layers are denoted by $\mathbf{h}_{i,j}^{(t,l)}$. This notation represents the vectorized input of a patch centered on location $(i,j)$ in layer $l$ computed at time step $t$ across all feature maps indexed by $k$. Thus an input image presented to the network is denoted as $\mathbf{h}_{i,j}^{(t,0)}$. 
The activations $z$ of a hidden recurrent layer then become
\begin{equation*}
	z^{(t,l)}_{i,j,k} =\left( \mathbf{w}_{k}^{(l)B} \right)^\top \mathbf{h}^{(t,l-1)}_{i,j} + \left( \mathbf{w}_{k}^{(l)L} \right)^\top \mathbf{h}^{(t-1,l)}_{i,j} + \left( \mathbf{w}_{k}^{(l)T} \right)^\top \mathbf{h}^{(t-1,l+1)}_{i,j},
\end{equation*}
where $\mathbf{w}_{k}^{(l)\cdot}$ is the vectorized convolutional kernel at feature map $k$ in layer $l$ for bottom-up (B), lateral (L), and top-down (T) connections, respectively. Each of these kernels is only active for models using the particular connection-type and is otherwise set to zero. As lateral and top-down connections depend on values of the previous time step, we define their inputs to be a vector of zeroes for $t=0$.
The $z^{(t,l)}_{i,j,k}$ of the hidden layer are batch-normalized \cite{ioffe2015batch}, passed on to rectified linear units (ReLU, $\sigma_z$), and go through local response normalization (LRN, $\omega$)  \cite{krizhevsky2012imagenet}, i.e.
\begin{equation*}
	h^{(t,l)}_{i,j,k} = \omega \left( \sigma_z \left( \mathrm{BN}_{\mathbf{\gamma},\mathbf{\beta}} \left( z^{(t,l)}_{i,j,k}\right) \right) \right).
\end{equation*}
After the second hidden layer the activations are relayed to a fully-connected layer with one output-unit per class and a softmax activation function.

\paragraph{Learning}
To quantify the mismatch between the networks' output $\hat{\mathbf{y}}^{(0,\dotsc,\tau-1)}$ and the one-hot target label $\mathbf{y}$ we compute the cross-entropy summed across all $\tau$ time steps and all $N$ output units:
\begin{equation*}
	J(\hat{\mathbf{y}}^{(0,\dotsc,\tau-1)}, \mathbf{y}) = - \sum_{t=0}^{\tau-1} \sum_{i=0}^{N} y_i \cdot \log \hat{y}^{(t)}_i + (1-y_i) \cdot \log(1- \hat{y}^{(t)}_i).
\end{equation*}
The network parameters are adapted using adam \cite{kingma2014adam} with an initial learning rate of $\eta = 0.003$. Training occurred for 25 epochs with mini-batches of size 500. Bottom-up weights were initialized with a truncated normal distribution with $\mu = 0$, $\sigma = \nicefrac{2}{\textrm{kernelsize}}$, all other weights with $\mu = 0$, $\sigma = 0.1$.

The different models were evaluated in terms of classification accuracy averaged across the test set. We use pair-wise McNemar's tests to compare test performances \cite{dietterich1998approximate}. To mitigate the increased risk of false positives, we control the false discovery rate (FDR) at 0.05 using a Bonferroni-type correction procedure developed in \cite{benjamini1995controlling}.
\section{Results}
\begin{figure}[hbtp]
\centering
\includegraphics[width=0.24\textwidth]{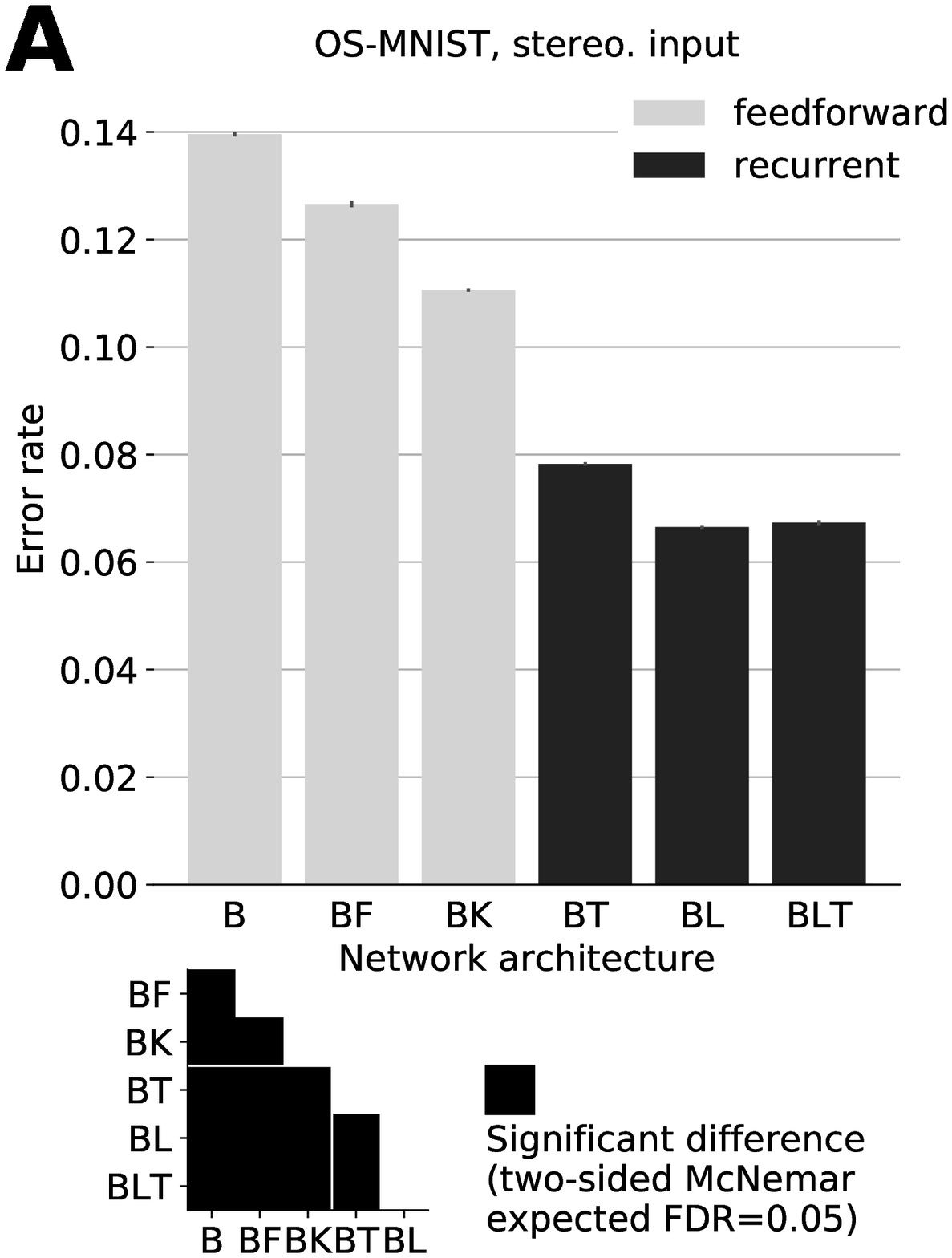}
\includegraphics[width=0.24\textwidth]{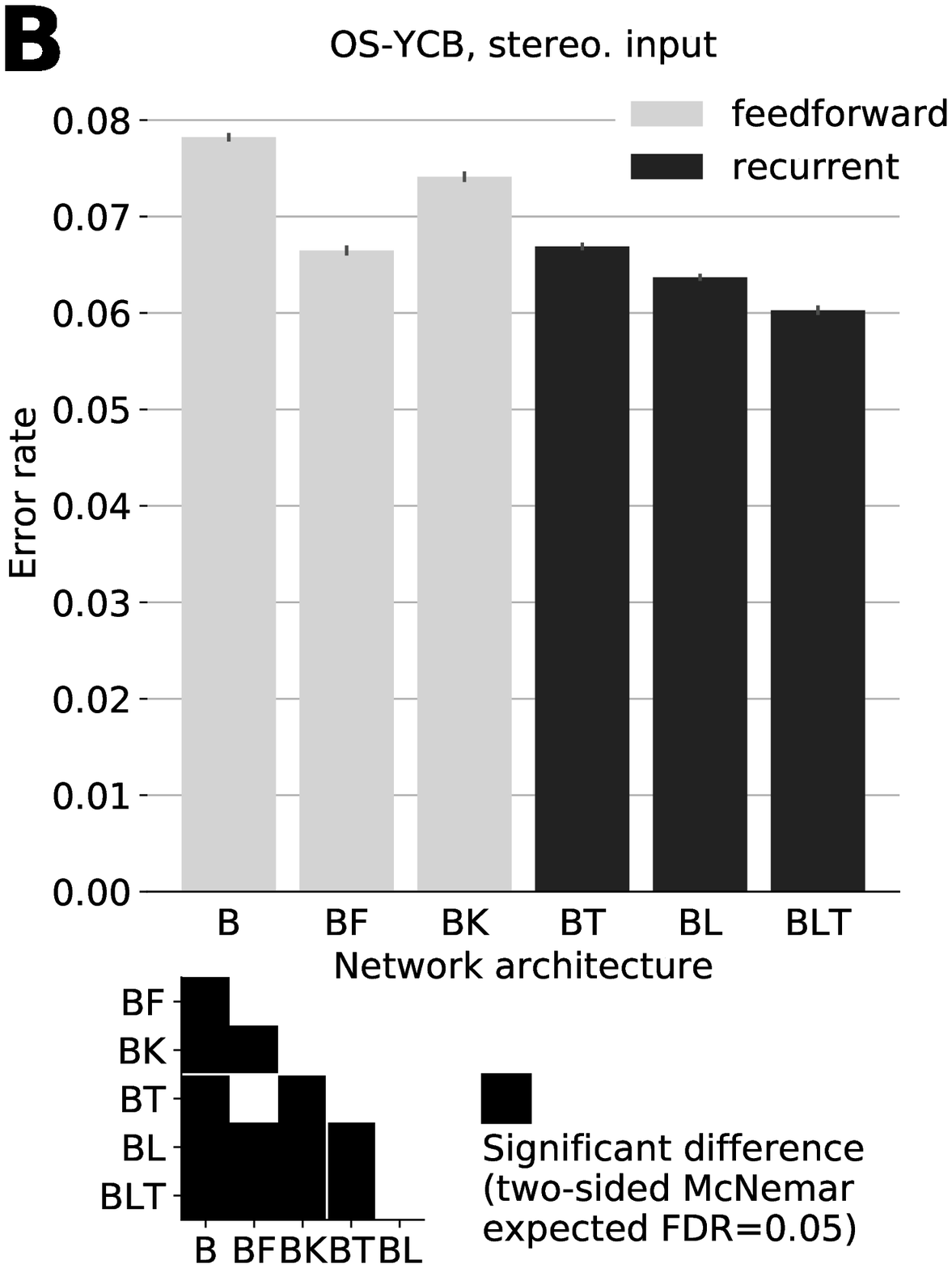}
\includegraphics[width=0.48\textwidth]{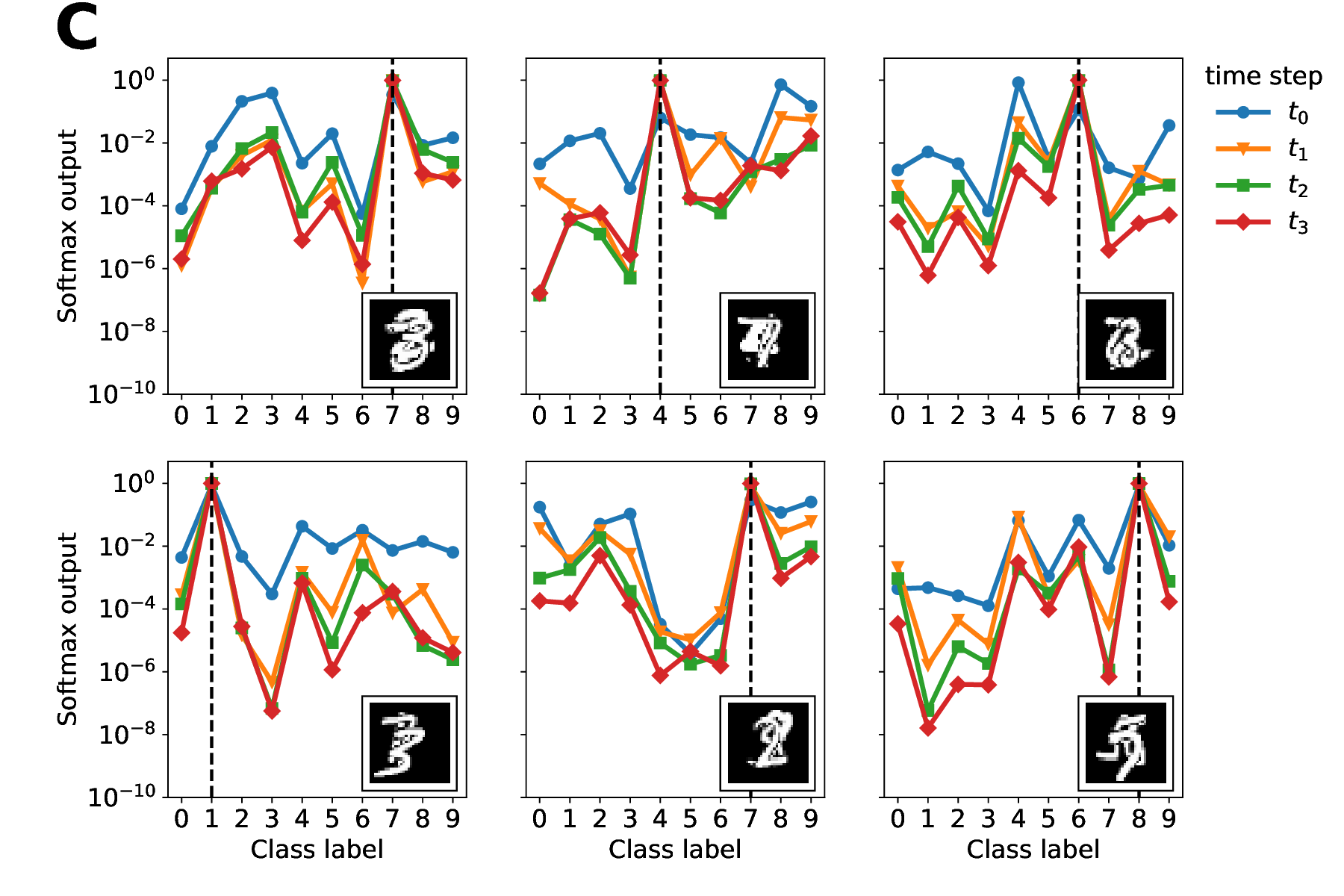}
\caption{Performance of the six network models, stereo input. Error bars indicate the standard error based on five repetitions of the training. Matrices depict results of pairwise McNemar tests. (A) OS-MNIST. (B) OS-YCB. (C) Softmax output (\emph{BLT}) over time, illustrating the effect of recurrent feedback. Correct label is indicated by dashed line.}
\label{fig:performance_results}
\end{figure}
Overall, recurrent architectures perform better than feedforward networks of near-equal complexity. Fig.~\ref{fig:performance_results} depicts the error-rate for the models trained with stereo input (A, B).
The lower left $3 \times 3$ squares of Fig.~\ref{fig:performance_results}, highlighted by a white line, indicate that all but one pair-wise test between feedforward and recurrent models show a significant performance gain for recurrent architectures. Only for OS-YCB (Fig.~\ref{fig:performance_results} B), \emph{BT} does not significantly outperform \emph{B-F}, $\chi^2(1,N = 63{,}200) = 0.52, p = .42$. For OS-MNIST significant differences (FDR = 0.05) can be attested for every combination except (\emph{BL}, \emph{BLT}).

Regarding the feedforward models, OS-MNIST shows a significant advantage for \emph{B-K} over \emph{B-F}. This is in contrast to OS-YCB, where the opposite holds true, $\chi^2(1,N = 63{,}200) = 16.41, p < .01$. This might be due to a larger number of specialized kernels better representing the larger amount of classes. Amongst recurrent models \emph{BT} performs worst and \emph{BLT} best in all runs. 

Qualitatively, monocular results resemble the discussed findings for stereoscopic input, see Tab.~1. 
However, the relative performance advantage of recurrent models is amplified for the stereoscopic case in both datasets. 

While OS-YCB has 79 possible classes, OS-MNIST provides only 10. The latter, however, has more in-class variability and, for monocular input, provides no cue but the partial visibility as to where the target object is located. The error rates for OS-MNIST (mono., range: .397 -- .539) and OS-YCB (mono., range: .178 -- .209) reflect these characteristics.


The softmax output indicates how confident the network ``feels'' about each class being the target. We observe that for more than \nicefrac{1}{10} of the  stimuli (OS-MNIST, stereo.) wrong initial guesses at $t_0$ are corrected at later time steps. Correct initial guesses are generally reinforced, only $\sim$$1.9\%$ of correct initial guesses are reverted, for specific examples see Fig.~\ref{fig:performance_results} C. 

\begin{table}[hbt]
\setlength{\tabcolsep}{5pt} 
\scriptsize
\centering
\begin{tabular}{cccccccc}


\toprule
OS-MNIST & \emph{B} & \emph{B-F}  & \emph{B-K} & \emph{BT} & \emph{BL} &  \emph{BLT} \\
\midrule

\multirow{1}{*}{Mono}
& $.539 \pm .001$ & $.507 \pm .001$      & $.495 \pm .001$    &  $.438 \pm .002$ & $.405 \pm .001$ & $\mathbf{.397 \pm .000}$  \\
\multirow{1}{*}{Stereo}
& $.140 \pm .000$ & $.127 \pm .001$      & $.111 \pm .000$    &  $.078 \pm .000$ & $\mathbf{.067 \pm .000}$ & $\mathbf{.067 \pm .000}$  \\
\\

\midrule
OS-YCB & \emph{B} & \emph{B-F}  & \emph{B-K} & \emph{BT} & \emph{BL} &  \emph{BLT} \\
\midrule

\multirow{1}{*}{Mono}
& $.209 \pm .001$ & $.196 \pm .000$      & $.205 \pm .001$    &  $.188 \pm .000$ & $.187 \pm .001$ & $\mathbf{.178 \pm .000}$  \\
\multirow{1}{*}{Stereo}
& $.078 \pm .000$ & $.066 \pm .001$      & $.074 \pm .001$    &  $.067 \pm .000$ & $.064 \pm .000$ & $\mathbf{.060 \pm .000}$  \\
\bottomrule
\end{tabular}

\label{tab:performance_comparison_sdd}
\caption{Error rates for all models, standard error based on five repetitions of training. Best performance per dataset is highlighted in bold.}
\end{table}

\section{Discussion}
We investigated whether recurrent connectivity benefits occluded object recognition. Previous attempts at answering this question have been limited by very simplistic and unnatural stimuli. On the one hand, the stimuli used by \cite{spoerer2017recurrent} were computer rendered digits without any variability in individual digit appearance. On the other hand, the stimuli used by \cite{oreilly2013recurrent, tang2014spatiotemporal} only blurred out image parts rather than introducing occluding objects. To overcome these limitations, we introduced two novel datasets that capture the natural variability of object appearance and a range of disparity and perspective cues. We demonstrated that feedback connections significantly improve occluded object recognition for these more complex datasets, providing strong evidence for a general benefit of recurrence for occluded object recognition.

In our experiments, the recurrent model with both lateral and top-down connections (\emph{BLT}) performed best in all runs. The \emph{BL} model came in second, while \emph{BT} performed worst, suggesting that lateral connections are particularly important for the observed performance gains.
A second finding is that recognition rates were higher for stereoscopic input. This is likely due to the second perspective of the scene, potentially revealing additional information about the target. Moreover, stereoscopic input provides the network with another cue regarding which parts of the input can be safely ignored: Only the target is presented at zero disparity, while the occluders are not.

Qualitatively, the results of the statistical network comparisons for monocular and stereoscopic inputs resemble each other. Interestingly, however, the relative performance difference between recurrent and feedforward models was usually higher for stereoscopic stimuli. This suggests that the recurrent connections are effective in utilizing the additional cues provided by the binocular presentation of the scene.
During training, we consistently observed that the sum of recurrent weights (lateral and top-down) became slightly negative. We hypothesize that this bias might contribute to inhibiting occluders.

Finally, our analysis revealed that recurrent connections ``sharpen'' the output distribution of the network while often correcting wrong initial guesses after the first feedforward pass through the network. Analyses with larger and more complex network architectures are left for future work.

In conclusion, we have shown that recurrent neural network architectures show significant performance advantages for occluded object recognition. Given their improved performance and greater biological plausibility they deserve more thorough analysis.

\begin{footnotesize}


\bibliographystyle{unsrtshort}

\bibliography{ESANN_OA.bib}

\begin{thebibliography}{10}

\bibitem{dicarlo2012visualobject}
J.~J. DiCarlo, D.~Zoccolan, and N.~C. Rust.
\newblock How does the brain solve visual object recognition?
\newblock {\em Neuron}, 73(3):415--434, 2012.

\bibitem{potter1976shortterm}
M.~C. Potter.
\newblock Short-term conceptual memory for pictures.
\newblock {\em Journal of Experimental Psychology: Human Learning and Memory},
  2(5):509--522, 1976.

\bibitem{krizhevsky2012imagenet}
A.~Krizhevsky, I.~Sutskever, and G.~E. Hinton.
\newblock Imagenet classification with deep convolutional neural networks.
\newblock In {\em Advances in Neural Information Processing Systems}, pages
  1097--1105, 2012.

\bibitem{kar2019evidence}
K.~Kar, J.~Kubilius, K.~Schmidt, E.~B. Issa, and J.~J. DiCarlo.
\newblock Evidence that recurrent circuits are critical to the ventral stream's
  execution of core object recognition behavior.
\newblock {\em Nature Neuroscience}, 22(6):974--983, 2019.

\bibitem{johnson2005recognition}
J.~S. Johnson and B.~A. Olshausen.
\newblock The recognition of partially visible natural objects in the presence
  and absence of their occluders.
\newblock {\em Vision Research}, 45(25):3262--3276, 2005.

\bibitem{tang2014spatiotemporal}
H.~Tang, C.~Buia, R.~Madhavan, N.~E. Crone, J.~R. Madsen, W.~S. Anderson, and
  G.~Kreiman.
\newblock Spatiotemporal dynamics underlying object completion in human ventral
  visual cortex.
\newblock {\em Neuron}, 83(3):736--748, 2014.

\bibitem{spoerer2017recurrent}
C.~J. Spoerer, P.~McClure, and N.~Kriegeskorte.
\newblock Recurrent convolutional neural networks: A better model of biological
  object recognition.
\newblock {\em Frontiers in Psychology}, 8:1551, 2017.

\bibitem{oreilly2013recurrent}
R.~C. O'Reilly, D.~Wyatte, S.~Herd, B.~Mingus, and D.~J. Jilk.
\newblock Recurrent processing during object recognition.
\newblock {\em Frontiers in Psychology}, 4:124, 2013.

\bibitem{ernst2019discounting}
M.~R. Ernst, J.~Triesch, and T.~Burwick.
\newblock Recurrent connections aid occluded object recognition by discounting
  occluders.
\newblock In I.~V. Tetko, V.~K{\r u}rkov{\'a}, P.~Karpov, and F.~Theis,
  editors, {\em Artificial Neural Networks and Machine Learning --ICANN 2019:
  Image Processing}, pages 294--305, Cham, 2019. Springer International
  Publishing.

\bibitem{lecun1998gradient}
Y.~LeCun, L.~Bottou, Y.~Bengio, and P.~Haffner.
\newblock Gradient-based learning applied to document recognition.
\newblock {\em Proceedings of the IEEE}, 86(11):2278--2324, 1998.

\bibitem{wyatte2012limits}
D.~Wyatte, T.~Curran, and R.~O'Reilly.
\newblock The limits of feedforward vision: recurrent processing promotes
  robust object recognition when objects are degraded.
\newblock {\em Journal of Cognitive Neuroscience}, 24(11):2248--2261, 2012.

\bibitem{calli2015benchmarking}
B.~Calli, A.~Walsman, A.~Singh, S.~Srinivasa, P.~Abbeel, and A.~M. Dollar.
\newblock Benchmarking in manipulation research: Using the yale-cmu-berkeley
  object and model set.
\newblock {\em IEEE Robotics \& Automation Magazine}, 22(3):36--52, 2015.

\bibitem{zeiler2010deconvolutional}
M.~D. Zeiler, D.~Krishnan, G.~W. Taylor, and R.~Fergus.
\newblock Deconvolutional networks.
\newblock In {\em 2010 IEEE Computer Society Conference on Computer Vision and
  Pattern Recognition (CVPR)}, pages 2528--2535. IEEE, 2010.

\bibitem{rumelhart1986learning}
D.~E. Rumelhart, G.~E. Hinton, and R.~J. Williams.
\newblock Learning representations by back-propagating errors.
\newblock {\em nature}, 323(6088):533, 1986.

\bibitem{ioffe2015batch}
S.~Ioffe and C.~Szegedy.
\newblock Batch normalization: Accelerating deep network training by reducing
  internal covariate shift.
\newblock In {\em International Conference on Machine Learning}, pages
  448--456, 2015.

\bibitem{kingma2014adam}
D.~P. Kingma and J.~L. Ba.
\newblock Adam: A method for stochastic optimization.
\newblock In {\em Proc. 3rd Int. Conf. Learn. Representations}, 2014.

\bibitem{dietterich1998approximate}
T.~G. Dietterich.
\newblock Approximate statistical tests for comparing supervised classification
  learning algorithms.
\newblock {\em Neural Computation}, 10(7):1895--1923, 1998.

\bibitem{benjamini1995controlling}
Y.~Benjamini and Y.~Hochberg.
\newblock Controlling the false discovery rate: A practical and powerful
  approach to multiple testing.
\newblock {\em Journal of the Royal Statistical Society. Series B
  (Methodological)}, 57(1):289--300, 1995.

\end{thebibliography}

\end{footnotesize}


\end{document}